%% file: aistats2024.tex
\documentclass[twoside]{article}

\usepackage{url}
\usepackage{graphicx}
\usepackage[accepted]{aistats2024}
\begin{document}

%

%

\twocolumn[

\aistatstitle{HintMiner: Automatic Question Hints Mining From Q\&A Web Posts with Language Model via Self-Supervised Learning}

\aistatsauthor{ Zhenyu Zhang \And JiuDong Yang }

\aistatsaddress{ Independent Researcher \And Independent Researcher}

]

\begin{abstract}
    Users often need ask questions and seek answers online.
    The Question - Answering (QA) forums such as Stack Overflow cannot always respond to the questions timely and properly. 
    In this paper, we propose HintMiner, a novel automatic question hints mining tool for users to help them find answers. 
    HintMiner leverages the machine comprehension and sequence generation techniques to automatically generate hints for users' questions.
    It firstly retrieve many web Q\&A posts and then extract some hints from the posts using MiningNet that is built via a language model.
    Using the huge amount of online Q\&A posts, we design a self-supervised objective to train the MiningNet that is a neural encoder-decoder model based on the transformer and copying mechanisms.
    We have evaluated HintMiner on 60,000 Stack Overflow questions. 
    The experiment results show that the proposed approach is effective. 
    For example, HintMiner achieves an average BLEU score of 36.17\% and an average ROUGE-2 score of 36.29\%. 
    Our tool and experimental data are publicly available. \footnote{\url{https://github.com/zhangzhenyu13/HintMiner}}.
\end{abstract}

\input{introduction.tex}
\input{background.tex}
\input{answer.tex}

\input{evaluation.tex}

\input{related.tex}
\input{conclusion.tex}

\bibliography{ref}

\end{document}

%% file: introduction.tex
\section{Introduction}

It is a common practice to seek answers from online Question and Answering (Q\&A) forums, such as Stack Overflow, Data Science, etc. \cite{wang2018understanding,chen2018data,calefato2018ask}. 
These Q\&A forums store abundant question related posts accumulated over years. 
However, as the posted questions in Q\&A sites rely on community members' voluntary answers, there is no guarantee to obtain timely and satisfactory answers for everyone question. 
As a matter of fact, we have found that a large number of questions lack accepted answers in Stack Exchange. 
It also costs users lots of time to search from those webs, where 
the Q\&A resource aggregating and reforming methods are quite a necessity.

In recent years, some methods have been proposed to help users with Q\&A. 
Some retrieval based methods such as AnswerBot \cite{xu2017answerbot} or the official Stack-Overflow website specify the key points of answers from the retrieved relevant posts by selecting the most important paragraphs.
Another kind effective Q\&A method is machine reading comprehension (MRC), which aims to understand the semantics of question and then select a text span from a given passage \cite{rajpurkar2018know,wang2018multi,wang2017r-net,chen2017reading} as the answer to the question.
However, the MRC cannot combine several spans to form a more rich and semantic-complete result. 
Enlightened by the Q\&A systems based on retrieval and MRC such as DrQA\cite{chen2017reading}, etc., 
we build a dedicated automatic question hints mining system to help users.
We targeted at mining hints from Q\&A forums while these methods do not utilize the specific Q\&A web resources.
And we also try to merge several selected spans to generate semantic rich and complete results while those previous works can only retrieve passages or select independent text spans.

In this paper, we aim to reuse the existing resources in online Q\&A forums to generate useful hints to the user questions. 
To that end, we propose a question hints mining tool called HintMiner, 
which selects and merges several useful segments of texts that can provide some hints for the question.
We formulate HintMiner as: \textbf{Find the most useful text spans from the relevant posts in Q\&A forums and combine them to generate the hints for the question.} 
Based on the hints provided, it will be much easier for users to get the final answers or help users to clarify and understand the questions.
HintMiner first leverages Elastic Search (\textbf{ES}\footnote{\url{https://www.elastic.co/elasticsearch/}}) to find relevant information for the question.
It then selects several text-spans that can provide some hints for a question to form the answer through machine reading comprehension \cite{chen2017reading}.
Finally, HintMiner merge the text-spans to generate semantic rich and complete hints via sequence generation \cite{ranzato2015sequence}.
To achieve this, we designed a self-supervised learning(SSL) objective for MiningNet to capture the semantics of questions and the relevant posts and to generate suitable hints. 
We construct "question" + "relative posts" + "proper hints/answers" triplets from millions of online stackoverflow posts.
Then we train the MiningNet to learn to generate such "hints/answers" with "questions" + "relative posts" as input.
MiningNet leverages BERT \cite{devlin2018bert} to encode the question and its relevant posts so as to capture their deep semantics.
The deep semantic representation is further fed to a transformer decoder \cite{vaswani2017attention} that can capture the importance of each input token through the attention mechanism.
With the learned token importance, we build a CopyNet using the copy mechanism \cite{gu2016incorporatingcopy2seq2seq,zhou2018sequentialcopy} to select a set of relevant tokens from the input to generate hints.

We have conducted extensive experiments to evaluate HintMiner. 
The results show that HintMiner outperforms several information retrieval based methods.
For example, HintMiner achieves an average of 36.17\% BLEU score and 36.29\% ROUGE-2 score. 
Furthermore, MiningNet outperforms several strong retrieval baselines and generation language model baselines.

Our contributions can be summarized as follows:
\begin{itemize}
    \item We build an automatic question hints mining tool called HintMiner, which can help developers solve questions.
    We extracted paragraphs from online Q\&A forums to build a useful posts dataset.
    We also make our code and data publicly available.
    \item We develop MiningNet, a novel self-supervised learning based model that can capture the semantics of a question and the relevant posts in Q\&A forums, and can generate the semantic rich and complete hints for questions.
    \item We have performed extensive evaluation of the proposed approach. 
    Our results show that HintMiner is effective and outperforms several strong baseline methods.
\end{itemize}

Our work is an important step towards intelligent hints mining for Q\&A forums.

%% file: background.tex
\section{The Q\&A Forums and Dataset}
\label{sec:bg}
\subsection{Question Answering Web Resources}

\begin{figure}[!t]
	\centering
	\includegraphics[width=2.50in]{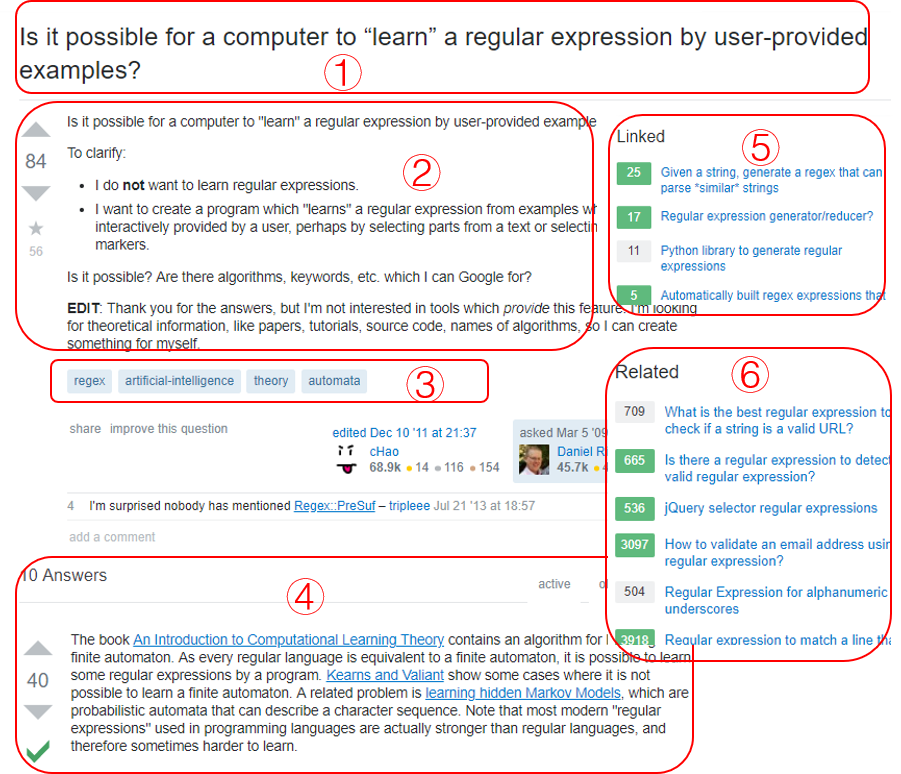}
	\caption{An example of the posts in Stack Overflow 
	}
	\label{fig-postpage}
\end{figure}

Web users would always ask questions or search relevant answers online.
To solve their problems, all kinds of online users depend heavily on online Q\&A sites. 
For example, Stack Overflow has become one of the most popular such Q\&A sites for developers, and it has accumulated a large number (over 16 million) of Q\&A posts. 
Figure \ref{fig-postpage} shows an example of the posts in Stack Overflow website. There are mainly six parts in a post: 1) the title of the question, showing the general concise description of the question, 2) the detailed description of the question, 3) the tags assigned by the user who posts the question, indicating the categories of the question, 4) the list of the answers to the question, including the accepted answer if available, 5) the linked posts that are marked by Stack Overflow community which are relevant to the current post, and 6) the related posts that are retrieved by the Stack Overflow system. 
It has been found that the responding time can be quite long and many questions may never be answered \cite{wang2018understanding}.  
It is desirable to improve question solution effectiveness by automating the question answering process.
Therefore, it is quite necessity to find proper hints for users' questions.

\subsection{The Construction of the Q\&A Dataset}
\label{sec:post-dataset}
\begin{figure}
	\centering
	\includegraphics[width=2.5in]{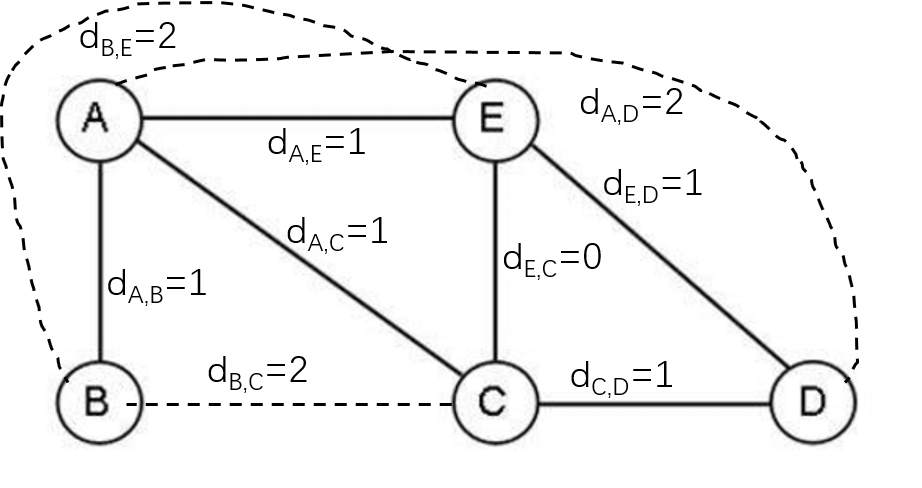}
	\caption{An Example of Post-Link Graph}
	\label{fig-link}
\end{figure}

We collected about 17 million posts from four Stack Exchange websites via \textit{\url{archive.org}}, including Stack Overflow\footnote{https://stackoverflow.com/}, Artificial Intelligence\footnote{https://ai.stackexchange.com/}, Data Science\footnote{https://datascience.stackexchange.com/} and Cross Validated\footnote{https://stats.stackexchange.com/}.
As illustrated in Figure \ref{fig-postpage}, the linked posts are marked by the community and are useful to the question.
There are about 19\% of the posts connected with over $5M$ links.
We build a Post-Link Graph where the nodes are posts and the edges are weighted links. There are two types of links marked by the community. 
We set the weights to 0 for links that mark duplicate posts and 1 for the others. 
Then we apply Dijkstra algorithm to compute the shortest link distance between each pair of nodes. 
Finally, we obtain 4 link distances (``0'', ``1'', ``2'' and ``$\geq$3'') because previous researches \cite{predictingsemantic,ye2017structure} show that two posts with a link distance $d \geq 3$ is not relevant to each other. 
For example, in Figure \ref{fig-link}, there are 6 marked links between $A,B,C,D,E$ and we complete the rest links (dashed lines) except for $B,D$ as $d_{B,D} \geq 3$. 
The link distance indicates how useful the post content is to the question of the other post.
The shorter the link distance is, the more useful the post is to the question.

\subsubsection{Selecting Relevant Posts For Training}
\label{sec:ret-train}
In order to train the MiningNet (Section \ref{sec:MiningNet}), we build a "question-passage-hints" triplets dataset.
For a question of the node (i.e. Post) in the post-link graph, we select top 2, 1, 1, 1 paragraphs for posts with distance as 0,1,2,and $\ge 3$ respectively to construct the relevant passage of current question.
The paragraphs of the passage are randomly shuffled so that the model cannot simply remember order of sentences.
We select the first passage of accepted answer in the post with more than 10 words as the \textbf{gold hints} for the question, which is considered to be meaningful.
We removed those questions without neighbors whose distance is 1.
Finally we constructed about 3.6 million "question-passage-hints" triplets.
Note that we add some less relevant paragraphs whose distance is larger than 1 so that noise and negative content are added to improve the robustness and difficulty of the dataset.
\label{sec:dataAnalysis}

\subsubsection{Selecting Relevant Posts For Inference}
\label{sec:ret-infer}

We first dumped the posts into the Elastic Search Engine (\textbf{ES}), and then retrieve relevant posts from a variety of Q\&A forums.
We then perform pre-processing of the selected posts.
In this work, we aim at generating hints rather than generating code or numerical expressions which usually exists in those scientific forums. 
Therefore, we replace a code snippet with \textbf{[CODE]}, and a mathematical expression with \textbf{[NUM]}. 
We do not consider hyperlinks either.
To reduce the vocabulary size, we use the BPE algorithm \cite{wordpiece} to perform tokenization, which can transform a compound word into a few tokens. 
For each question, we retrieve 5 posts in total.

The retrieved posts often contain many non-essential sentences that are useless and can make it difficult for a deep neural network to handle extremely long input \cite{6challenges-KoehnK17}. 
Examples of such sentences are "Maybe my answer can help you", "Thank you for your suggestion", etc. 
Therefore, we leverage an ensemble method to filter those sentence, which combines the results of three base algorithms that can identify the important sentences. 
1) \emph{Lexrank} \cite{lexrank}, a graph based method inspired by Pagerank algorithm\cite{pagerank}, which uses the eigenvector centrality of sentences to select the important sentences. 
2) \emph{KL greedy search} \cite{KL-sum}, an information entropy maximization based method, which uses the KL divergence to compute the relative information gain to greedily select sentences so as to maximize the information entropy of selected sentences. 
3) \emph{Latent Semantic Analysis (LSA)} \cite{LSA}, which decomposes the sentence-term matrix using SVD and selects the sentences with the most significant topics via the right singular vectors.
The three base algorithms focus on different aspects of sentence importance. 
Therefore, we merge their results and eliminate the sentence repetition. 
The resulting set of sentences forms the context passage for MiningNet. 

%% file: answer.tex
\section{HintMiner: Generating Hints to User's Questions}
\label{sec:answer}

\subsection{System Overview}
In our work, we formulate the problem as follows: given a question and a set of relevant posts, the core problem is to select a set of useful text spans from existing posts and generate the hints to the question. We process the posts to form the context passage for training (Section \ref{sec:ret-train}) and inference (Section \ref{sec:ret-infer}). 
The hints are then generated by the MiningNet.

\begin{figure}[!t]
	\centering
	\includegraphics[width=2.8in]{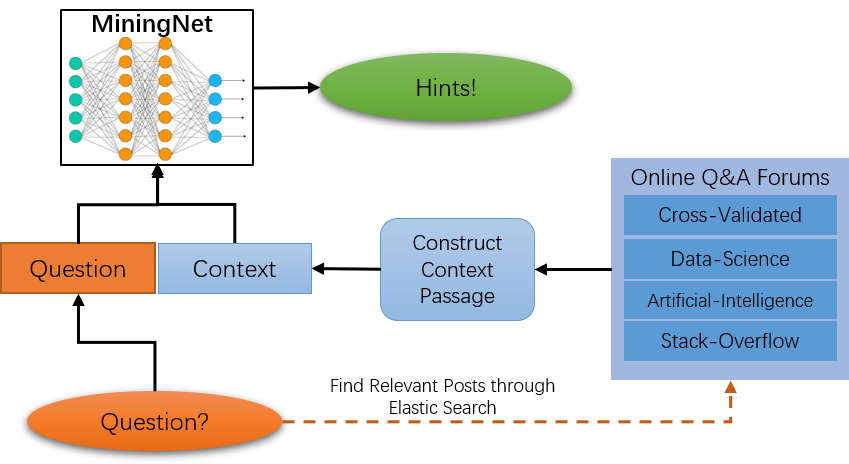}
	\caption{An Overview of HintMiner  
	}
	\label{fig-frameworks}
	\vspace{-10pt}
\end{figure}

For that purpose, we build HintMiner, which utilizes the techniques of machine reading comprehension \cite{chen2017reading} and sequence generation \cite{ranzato2015sequence}.
Figure \ref{fig-frameworks} shows the overview of HintMiner.
Given a question, we first select the relevant posts from the Q\&A forums to form the context passage (Section \ref{sec:ret-infer}). 
Then we feed the context and question to MiningNet, 
which is an effective deep neural network that can generate the hints to the question by copying and generating tokens from the context.

\subsection{The MiningNet Model}
\label{sec:MiningNet}

\begin{figure}[!t]
	\centering
	\includegraphics[width=3.0in]{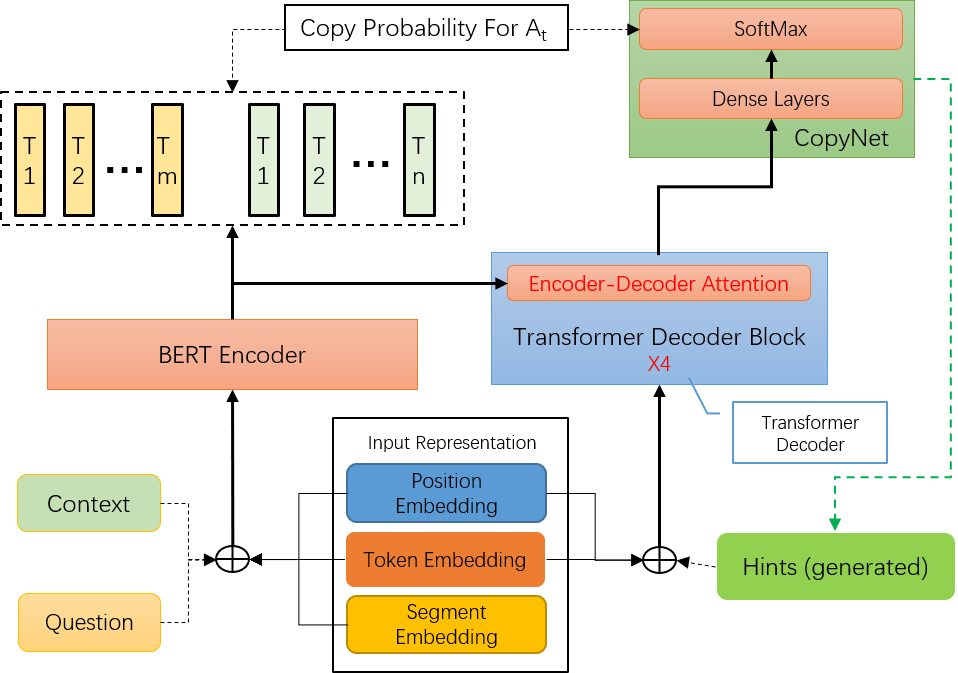}
	\caption{An Overview of MiningNet
	}
	\label{fig-MiningNet}
		\vspace{-10pt}
\end{figure}

\subsubsection{The Structure of the Model}

Figure \ref{fig-MiningNet} shows the structure of MiningNet, which consists of four parts: a BERT Encoder,
a transformer decoder, and a CopyNet. 
There are three embeddings in the Input Representation that takes Q\&A data as input and outputs the embeddings of input. 
These embeddings are extracted from BERT \cite{devlin2018bert}. 
It is worth mentioning that the segment id for tokens in questions is 0 and for tokens in context is 1.
When generating the $t^{th}$ answer token ($A_t$), the tokens of the generated answer before step $t$ ($A_1,...,A_{t-1}$)
are embedded using position embedding and token embedding only, and are then fed to the transformer decoder.
The $\oplus$ in Figure \ref{fig-MiningNet} refers to the use of BERT embedding layer to embed the tokens in text.
Equation \ref{eq-bert} presents how BERT is used to encode the sequence in our model. Each token in $q$ (a question) and $c$ (the context) is encoded as a $dim$ dimension dense vector $T_{i}^{q}$/$T_{j}^{c}$, and the $T_{cls}$ represents the pooling vector.
 \begin{equation}
 	\label{eq-bert} 
 	\begin{split}
 	  T=[T_{CLS},T_{1}^{q},,...,T_{m}^{q},T_{1}^{c},...,T_{n}^{c}]=BERT([q,c]]), \\
 	 where \, T_{i}^{q},T_{j}^{c} \in R^{dim}, \, 1 \leq i \leq m, \, 1 \leq j \leq n
 	\end{split}
\end{equation}

The output of the BERT encoder is a sequence of vectors representing the semantics of the question and context passage. 
The transformer decoder \cite{vaswani2017attention} reads the output of the BERT encoder and then computes the output hidden state of target (i.e. generated answer) vectors. 
The transformer decoder also computes the encoder-decoder attention score vectors, which uses the multi-head attention mechanism to pay attention to the ``question+context passage''. 
We also build a CopyNet \cite{gu2016incorporatingcopy2seq2seq,zhou2018sequentialcopy}, which takes the encoder-decoder attention vectors as input and outputs an answer text. 
Using CopyNet, certain text spans in the input sequence are selected to be present in the output sequence with the Copy Probability \cite{gu2016incorporatingcopy2seq2seq}. 
Thus, MiningNet can generate the hints to a question through the copying mechanism by selectively replicating the input text spans.  
In this way, we transform the hints generation problem to a MRC problem,  where the hints is composed of several selected text spans and combined via generation.
The hints tokens $A_1$, $A_2$, ... $A_n$ are generated one by one through the copying mechanism iteratively until $A_{n}$ is the end token or $n$ exceeds the hints length limit. 
(
The generation length ($n$) is usually set to a fixed length for satisfactory model performance  \cite{6challenges-KoehnK17}.
)

\subsubsection{Transformer Decoder and CopyNet}
 In this subsection, we describe the Transformer Decoder and CopyNet models in detail and show how to adapt them to MiningNet.
The encoder vectors ($T$) are the semantic representation of ``question+context''. We apply the transformer decoder to them and compute the encoder and decoder attention via Equation \ref{eq-decencattn}, which defines the compatibility function of the query with the corresponding key in the multi-head attention. 
$Q$ denotes the decoder hidden vectors at time step $t$, which is given as shifted masked $t-1$ true answer encoding vectors during training and $t-1$ predicted answer encoding vectors during testing. 

To select text spans from the context, 
we apply the copying mechanism  ($Copy$) on the encoder-decoder attention vectors, which can select tokens from input directly. 
According to the copy mechanism and the attention in Equation \ref{eq-decencattn}, the output of CopyNet (a three layer MLP with same model dimension and activation function used in BERT)
is denoted as $p(A_{t}|A_{1},...,A_{t-1},T)=Copy(attn^{t})=softmax(attn^{t})$, where $attn_{j}^{t}$ is the attention value at step $t$ for $j^{th}$ context vector.  
Therefore, the probability distribution for generated answer sequence $p(A_1),p(A_2),...,p(A_n)$ can be denoted as Equation \ref{eq-ansGen}.

\begin{equation}
 \label{eq-decencattn} 
 \begin{split}
    Attention(Q,K,V)=softmax(\frac{Q*K^{T}}{\sqrt{dim}})*V,\\
    where \, K=V=BERT([q,c])
 \end{split}
\end{equation}

\begin{equation}
 	\label{eq-ansGen} 
 	p(A|T)=\prod_{ t }p(A_{t}|A_{1},...,A_{t-1},T)
\end{equation}

It is worth mentioning that the hints generation process is based on captured semantics of question and context (passage formed from relevant posts).  
Essentially, MiningNet simulates a function that maps the semantic representation of the \textbf{context to hints} text spans according to the \textbf{requirement of the question} via the attention mechanism. 
The context contains the knowledge that can provide some hints for the question, which is represented as encoded vectors. 
The BERT encoder leverages its well designed and pre-trained network to represent the semantics of the question and context, then the decoder computes an attention score for each token of hints that is to be copied from the context based on the semantics. 
In the end, the selected text-spans could represent the most proper hints that can help to clarify and understand the question.

\subsection{Self-supervised Learning Objective}

We aim to \textbf{train the MiningNet to learn to specify the most useful text-spans from a given context paragraph for the question.}
Therefore, we propose our SSL objective: let the model learn to distinguish important sentences, phrases or words as hints from a noisy context input given a question.
For each post, we extract the top rated K(=3) answers and shuffle them randomly to prevent the model remembering to copy the best one always.
We concatenate the K answers and feed the resulting text to the unimportant sentences filtering component to form the context.
We use the best answer as the (the most useful) answer to the question of the post.
Finally we form over 1,900,000 $<question, context, hints>$ triplets for training the model.

\subsection{The Implementation and Training Details}
\label{sec:implementation}

HintMiner utilizes the MiningNet to understand the semantics of questions and posts, and then generate suitable answers to the questions. To implement MiningNet, we leverage the transformers library \cite{wolf-etal-2020-transformers}. 
We use the BERT-base as backbone. 
For encoder-decoder attention we use 12 attention heads and the hidden size is the same as the BERT.
We set hyper-parameters based on previous research\cite{massiveExplore-Britz:2017} and the pre-trained BERT encoder structure\cite{devlin2018bert}.
Therefore, the hyper-parameters are well fine-tuned.

As the encoder-decoder model suffers from the exposure bias issue \cite{ranzato2015sequence,yuan2017machine,he2016dual}, we adopt a hybrid training strategy which firstly uses teacher forcing training \cite{ranzato2015sequence} in a supervised way and then uses the policy gradient reinforcement learning with BLEU4 \cite{BLEU} as reward to fine-tune the model. 
We leverage the Adam optimizer \cite{deeplearning-book} to maximize the probability denoted in Equation \ref{eq-ansGen}. 
We train the model with initial learning rate as $1e-5$ for 200,000 training steps with batch size as 32.

%% file: evaluation.tex
\section{Experiments}
\label{sec:exp}

\begin{table*} [!t]
	\caption{Evaluation of HintMiner and Compared Methods. ``ROU-2'' denotes ROUGE-2.}
	\label{table-IR}
	\centering
	\begin{tabular}{l|ll|ll|ll|ll}
		\hline
		\multicolumn{1}{c|}{} &
		\multicolumn{2}{c|}{50} &
		\multicolumn{2}{c|}{100} &
		\multicolumn{2}{c|}{150} &
		\multicolumn{2}{c}{200} \\
		\cline{2-9}
		\multicolumn{1}{c|}{} 
		& \bfseries \boldmath BLEU & \bfseries \boldmath ROU-2 
		& \bfseries \boldmath BLEU & \bfseries \boldmath ROU-2 
		& \bfseries \boldmath BLEU & \bfseries \boldmath ROU-2 
		& \bfseries \boldmath BLEU & \bfseries \boldmath ROU-2 
		\\
		\hline
		AnswerBot & 14.22 & 16.81 & 15.63 & 19.63 & 17.58 & 21.27 & 17.33 & 21.28 \\
		SimCSE & 16.53 & 18.05 & 16.72 & 19.62 & 20.98 & 22.13 & 20.11 & 22.66 \\
		PageRank++ & 19.25 & 20.02 & 19.28 & 22.55 & 21.84 & 22.76 & 19.67 & 22.84 \\
		\hline
        GPT2 & 25.11 & 26.13 & 26.49 & 26.83 & 26.53 & 28.51 & 28.04 & 29.16 \\
  	BART & 29.35 & 29.57 & 30.75 & 31.29 & 31.52 &         32.45 & 33.11 & 33.17 \\
		UniLM & 30.00 & 31.16 & 33.25 & 35.17 & 34.88 & 33.02 & 33.69 & 33.84 \\
        \hline
		HintMiner  & 32.01 & 32.28 & 36.17 & 36.29 & 36.09 & 36.13 & 34.32 & 34.67 \\
        \hline
        
	\end{tabular}
\end{table*}

\begin{table}[!t]
    \renewcommand{\arraystretch}{1.3}
    \caption{Comparison of HintMiner with Different Relevant Post Retrieval Methods}
    \label{table-retriever}
    \centering
    \begin{tabular}{ccc}
        \hline
        \bfseries \boldmath Retrieval Methods &  \bfseries \boldmath BLEU & \bfseries \boldmath ROUGE-2 \\
        \hline
        Linked Posts (User Marked) & 36.63 & 36.85 \\
        Stack Exchange open API & 36.22 & 36.25 \\
        Google Custom Search & 36.19 & 36.37 \\
        Elastic Search (HintMiner) & 36.17 & 36.29 \\
        \hline
    \end{tabular}
        	\vspace{-10pt}
\end{table}

\begin{table*}[!t]
    \caption{Examples of Hints Generated by HintMiner}
    \label{table-exmaples}
    \centering
    \normalsize
    \begin{tabular}{p{6cm}|p{9cm}}
        \hline
        \bfseries \boldmath Question &  \bfseries \boldmath Hints \\

        \hline
        
        how do I modify an existing a sheet in an excel workbook using openxlsx package in r? (https://stackoverflow.com/
        questions/34172353)
        &
        you need to load the complete workbook, then modify its data and then save it to disk. with [CODE] you can also specify the starting row and column. and you could also modify other sections. 
        \\
        \hline
        how do you save android emulator snapshot?
         (https://stackoverflow.com/
         questions/4842612/)
         &
         start a telnet session to the android emulator then freezes for a few seconds while saving/loading a snaps. i found at google, simply closing the emhot window is the correct way to you.
         \\
        \hline
        
        is there a method to calculate something like general similarity score of a string?
        (https://stackoverflow.com/
        questions/4323977/)
        &
        there are many such algorithms. keywords are fuzzy string matching. by it you can calculate the number of changes required to transform one string into another, so that gives you an estimate of how similar the strings are.
        \\
        \hline
    \end{tabular}
        
\end{table*}

\subsection{Experimental Design}
We conducted experiments to evaluate the effectiveness of HintMiner. 
Our evaluation focuses on the following four research questions:

\textbf{RQ1: How effective is HintMiner in generating hints?} 

We compare our model with two kinds of representative methods, i.e. the text retrieval and text generation techniques. 
We list the compared methods as follows:
\begin{itemize}
    \item  \textbf{Text retrieval based methods}.
    \textbf{AnswerBot} \cite{xu2017answerbot} applies the MMR algorithom \cite{carbonell1998use} to the relevant posts given a question to extract proper paragraphs as answers.  
    \textbf{SimCSE} is well trained via contrastive learning and can behave rather well in specifying semantic relevant sentences \cite{simcse21, ConSERT21}.
    We retrieve the the sentences with highest cosine similarity for a question from the posts. 
    \textbf{PageRank++} extends the traditional graph-based important sentences selection approaches \cite{lexrank,textrank} by replacing the tf-idf features of PageRank\cite{pagerank} with semantic vectors of SimCSE.

    \item  \textbf{Text generation based methods}.
     \textbf{GPT2} \cite{radford2019language} an auto-regressive language model that is pre-trained to predict the next token in text, which is good at many text generation tasks such as summarization, answer generation, etc.
     \textbf{BART} \cite{raffel2020exploringT5} leverages the advantages of both BERT \cite{devlin2018bert} and GPT models \cite{radford2019language} to pretrain a encoder-decoder based language models by applying several language mask strategies in tokens, sentences and whole documents.
     \textbf{UniLM} \cite{bao2020unilmv2} is a pre-trained unified language model for both auto-encoding and partially auto-regressive language modeling tasks using a a pseudo-masked language model, which is good at language understanding and generation.

    For each method in baselines, we apply same data processing methods as our HintMiner to prompt fair
    comparable results. 
    In our implementation, we leveraged the huggingface transformers \cite{wolf-etal-2020-transformers} to build the neural networks. 

\end{itemize}

\vspace{3pt}
\textbf{RQ2: How effective is HintMiner when different relevant post retrieval methods are used?}

The relevant post retrieval is an important part of HintMiner. 
In HintMiner, we use Elastic Search Engine to search for relevant posts. 
In this RQ, we evaluate the influence of different retrieval methods.
In Q\&A forums such as Stack Overflow, community members often manually mark the linked posts for some questions. 
We experimented with the linked posts as the relevant posts (i.e. we directly select posts based on the post-link graph as described for training in Section \ref{sec:ret-train} ).  
Refer to Section \ref{sec:post-dataset} for more details.
We also leverage the open online methods such as Stack Exchange Search Engine API\footnote{\url{https://api.stackexchange.com}} and Google Search Engine API \footnote{\url{https://developers.google.com/custom-search}} to retrieve three related posts for each test post.

\vspace{3pt}
\textit{\textbf{Experimental settings:}}
To evaluate the effectiveness of HintMiner, we randomly sampled 60,000 posts accepted answers that do not appear in our training data. 
The first paragraph of accepted answer with more than 10 words are \textbf{gold hints} of the question.
Then, for each question in the posts, we used HintMiner to generate the hints. 
Finally we used 4-gram BLEU score and 2-gram ROUGE score (ROUGE-2) to evaluate the quality of the generated hints. 
For RQ1, RQ2 and RQ4, we set the context length to 500 words in our experiments. 
In the generation decoding process, we use the BEAM-Search algorithm with beam size as 5 and select the best generated texts as the final hint for a question.

\subsection{Evaluation Metrics}
 
To evaluate the generated hints, we use two n-gram language model evaluation metrics, i.e. ROUGE and BLEU \cite{ROUGE,BLEU}, which are widely used in machine translation, summarization, and text generation tasks, etc., to measure the similarity between two sentences. 
In our research, we measure whether the generated hints are similar to the gold hints.
The BLEU score uses the common presence of n-gram count of generated text and reference text to measure the similarity from the precision-like perspective. 
The ROUGE score measures the similarity from the recall-like perspective. 
In our experiment, ROUGE uses 2-gram (i.e. ROUGE-2) and BLEU uses 4-gram.
Both BLEU and ROUGE-2 scores are 100\% when the generated hints are the same as the true hints and 0 when they are totally different. 
The larger the value, the better the generated hints will be.

\subsection{Experimental Results for RQ1}

The three retrieval baselines are given in Table \ref{table-IR} for the first three rows.
It shows the performance results of all the experimented methods for different answer lengths respectively,  
where $50,100,150,200$ are the answer tokens we truncated. 
The SimCSE outperforms the AnswerBot (that is based on hand-crafted features) a lot which demonstrate that the capturing the semantics of question and passage through BERT is critical.
The BERT++ outperforms the SimCSE, which demonstrates that the importance of sentences in paragraphs concerns a lot and thus it's natural to apply some attention mechanism to better select proper sentences.
The HintMiner here directly select text-spans with rather than coarse sentence-level granularity, which significantly outperforms all the baselines.

For the generation based baselines,
as shown with middle three rows in Table \ref{table-IR}, HintMiner significantly outperforms the three baselines using the proposed MiningNet.
The MiningNet obtains a BLEU score of 36.17\% and a ROUGE-2 score of 36.29\% when the maximum answer length is set to 100 words. 
The HintMiner outperforms all the baselines given the length of the generated answers varies from 50 to 200 words, which shows the effectiveness of the proposed pre-training objectives.
The public models such as GPT2, BART and UniLM is trained using common language modeling objectives, which is not a good solution for the professional situations in our research.
Through the unsupervised learning with the huge amount of programming posts, the MiningNet is able to select the needed text spans from the context to generate answers that are semantically similar to the true answers.

We manually checked about 100 questions and hints generated from those methods.
The results showed that the sentences that are similar to the question are not necessary the sentences that can form the hints that need to useful for the question rather than just repeat the meaning of question again. 
Also, the hints are not necessary to be one complete sentence because some contents in paragraphs are not necessary.
Therefore, selecting from sentence-level is not rational (i.e. the 3 retrieval baselines).
It is also sub-optimal to only consider the common language semantics from Wikipedia or Bookcorpus to pre-train a language model, which lack of knowledge for a specific domains and are not trained to distinguish useful contents as hints in our research problem.
In conclusion, these baseline methods cannot effectively generate proper hints given noisy relative posts, which leads to lower performance.

\subsection{Experimental Result for RQ2}

Table \ref{table-retriever} shows the effectiveness of HintMiner when using different methods to find the relevant posts. 
Using the Linked Posts manually marked by the Q\&A community, HintMiner can achieve the best performance, but there are only around 19\% of posts marked with links and newly posted questions lack these user marked links. 
When using Google Custom Search, the search service by Stack Exchange and Elastic Search, both BLEU score and ROUGE-2 score drop a little, but the performance is still acceptable. 
This experiment also demonstrates the stability and scalability of HintMiner for dealing with different sources retrieved as context passage.

\subsection{Examples of the Generated Hints}

Table \ref{table-exmaples} shows some of the hints generated by HintMiner, where the $1^{st}$ is a solved question (i.e., questions with accepted answers) and the rest are not solved yet. 
We omit the detailed description of questions and provide the link to the corresponding Stack Overflow page. 
The HintMiner can propoerly select text spans of accepted answers against noise paragraphs of sentences (Section \ref{sec:post-dataset}).
Although the results may contain grammatical errors they are generally readable and useful.  
Currently, the generated hints do not contain code or mathematical expression (i.e. represented with symbols such as \textbf{[NUM]} and \textbf{[CODE]}). 
For example, the accepted answer of the $2^{nd}$ question is attached with a code snippet while our generated hints just shows the presence of code here (\textbf{[CODE]}).

To further evaluate the effectiveness of HintMiner, we also randomly sampled some without accepted answers. 
We can see that HintMiner is able to generate meaningful and useful hints even without gold hints in the passage.
Taking the $2^{nd}$ hint as an example, the question is about "usage of simulator" and the answer provides some tips for the question. 
Those answers further confirm the usefulness of HintMiner. 
These results are encouraging.

%% file: related.tex
\section{Related Work}
\label{sec:rw}

In recent years, question answering (Q\&A) has been receiving a lot of attention in natural language processing.
Generally, there are mainly three kinds Q\&A systems, including IR based Q\&A, KBase based Q\&A, and MRC based Q\&A. 
Antonio et al. \cite{soares2018literature} conducted a literature review in the 130 out of 1842 papers on Q\&A systems, and found that 28.57\% of the surveyed papers are based on IR and 34.9\% on KBase. 
IR \cite{croft2010search} is widely studied in Q\&A, and combining its with KBases based methods to
fetch answers by searching knowledge bases \cite{dong2015question, yih2015semantic} are gaining momentum as some established knowledge bases like FreeBase and DBpedia are publicly available. 
Recently, many MRC based Q\&A methods \cite{chen2017reading,wang2018multi,wang2017r-net} have been proposed. 
For example, Miller et al. \cite{miller2016key} used MRC on wikipedia to find the text spans for questions. 
Currently, these research mainly focus on general open domain Q\&A and lack support for the utilization of Q\&A forums resources. 
To help better understand HintMiner, we introduce the MRC and Copy Mechanism here briefly.

\subsection{Copying Mechanism}

Copying mechanism is inspired by pointer network \cite{vinyals2015pointer} that is proposed for OOV problems. 
It is widely used in many Seq2Seq models \cite{gu2016incorporatingcopy2seq2seq,zhou2018sequentialcopy}.
The copy mechanism selects a set of input tokens to the output. 
The CopyNet based on the pointer-network \cite{vinyals2015pointer, zhou2018sequentialcopy} directly leverages the attention between the source encoder and the target decoder to generate an output probability distribution over the source tokens.

\subsection{Machine Reading Comprehension}
Machine reading comprehension (MRC) comprehends a natural language question and then selects a text span (usually not longer than 40 tokens) from a given passage \cite{rajpurkar2018know,wang2018multi,wang2017r-net,chen2017reading} as the answer to the question. 
For each question, the task is to select a text span to answer it by outputting a start index and an end index of the input sequence tokens of the passage. 
For example, DrQA \cite{chen2017reading} select spans over millions of  Wikipedia pages to answer general questions such as ``who is the current president of USA?". 
Wang et al. proposed a multi-granularity attention fusion networks \cite{wang2018multi} to encode the question and the article via multi-granularity attention to select proper text-span. 
Microsoft researchers also proposed R-Net \cite{wang2017r-net} to predict the answer text-span, which leverages the pointer-network for text-span selection. 

%% file: conclusion.tex
\section{Conclusion}
\label{sec:con}
In this paper, we have proposed HintMiner, a machine comprehension and generation based approach to automatic mining hints for users' questions. 
Given a new question, HintMiner first selects relevant posts and filters away unimportant sentences in the retrieved posts from ES. 
It then utilizes MiningNet to generate hints to the question from the paragraphs of the relevant posts. 
MiningNet is an effective self-supervised learning based model, which is able to distinguish proper contents from relevant posts to generate hints. 
We conduct extensive experiments to evaluate the model effectiveness. 
The evaluation results show that HintMiner outperforms several important Q\&A methods and MiningNet is an effective hints generation neural network. 
Our tool and experimental data are publicly available \url{https://github.com/AnonymousAuthor2013/HintMiner}.

In the future, we plan to build a link prediction tool that can better find more relevant posts for a given question. 
To further improve the capacity of HintMiner, we will also investigate models to comprehend code(\textbf{[CODE]}) and numerical expressions (\textbf{[NUM]})in posts. 
We will also explore more effective metric and perform user studies to evaluate the usefulness of our tool in practice.

%% file: ref.bib
@article{soares2018literature,
  title={A Literature Review on Question Answering Techniques, Paradigms and Systems},
  author={Soares, Marco Antonio Calijorne and Parreiras, Fernando Silva},
  journal={Journal of King Saud University-Computer and Information Sciences},
  year={2018},
  publisher={Elsevier}
}

@inproceedings{wolf-etal-2020-transformers,
    title = "Transformers: State-of-the-Art Natural Language Processing",
    author = "Thomas Wolf and Lysandre Debut and Victor Sanh and Julien Chaumond and Clement Delangue and Anthony Moi and Pierric Cistac and Tim Rault and Rémi Louf and Morgan Funtowicz and Joe Davison and Sam Shleifer and Patrick von Platen and Clara Ma and Yacine Jernite and Julien Plu and Canwen Xu and Teven Le Scao and Sylvain Gugger and Mariama Drame and Quentin Lhoest and Alexander M. Rush",
    booktitle = "Proceedings of the 2020 Conference on Empirical Methods in Natural Language Processing: System Demonstrations",
    month = oct,
    year = "2020",
    address = "Online",
    publisher = "Association for Computational Linguistics",
    url = "https://www.aclweb.org/anthology/2020.emnlp-demos.6",
    pages = "38--45"
}

@article{radford2019language,
  title={Language models are unsupervised multitask learners},
  author={Radford, Alec and Wu, Jeffrey and Child, Rewon and Luan, David and Amodei, Dario and Sutskever, Ilya and others},
  journal={OpenAI blog},
  volume={1},
  number={8},
  pages={9},
  year={2019}
}

@article{raffel2020exploringT5,
  title={Exploring the limits of transfer learning with a unified text-to-text transformer},
  author={Raffel, Colin and Shazeer, Noam and Roberts, Adam and Lee, Katherine and Narang, Sharan and Matena, Michael and Zhou, Yanqi and Li, Wei and Liu, Peter J},
  journal={The Journal of Machine Learning Research},
  volume={21},
  number={1},
  pages={5485--5551},
  year={2020},
  publisher={JMLRORG}
}

@inproceedings{bao2020unilmv2,
  title={Unilmv2: Pseudo-masked language models for unified language model pre-training},
  author={Bao, Hangbo and Dong, Li and Wei, Furu and Wang, Wenhui and Yang, Nan and Liu, Xiaodong and Wang, Yu and Gao, Jianfeng and Piao, Songhao and Zhou, Ming and others},
  booktitle={International conference on machine learning},
  pages={642--652},
  year={2020},
  organization={PMLR}
}

@article{simcse21,
  author    = {Tianyu Gao and
               Xingcheng Yao and
               Danqi Chen},
  title     = {SimCSE: Simple Contrastive Learning of Sentence Embeddings},
  journal   = {CoRR},
  volume    = {abs/2104.08821},
  year      = {2021},
  url       = {https://arxiv.org/abs/2104.08821},
  archivePrefix = {arXiv},
  eprint    = {2104.08821},
  bibsource = {dblp computer science bibliography, https://dblp.org}
}

@inproceedings{ConSERT21,
  author    = {Yuanmeng Yan and
               Rumei Li and
               Sirui Wang and
               Fuzheng Zhang and
               Wei Wu and
               Weiran Xu},
  editor    = {Chengqing Zong and
               Fei Xia and
               Wenjie Li and
               Roberto Navigli},
  title     = {ConSERT: {A} Contrastive Framework for Self-Supervised Sentence Representation
               Transfer},
  booktitle = {Proceedings of the 59th Annual Meeting of the Association for Computational
               Linguistics and the 11th International Joint Conference on Natural
               Language Processing, {ACL/IJCNLP} 2021, (Volume 1: Long Papers), Virtual
               Event, August 1-6, 2021},
  pages     = {5065--5075},
  publisher = {Association for Computational Linguistics},
  year      = {2021},
  url       = {https://doi.org/10.18653/v1/2021.acl-long.393},
  doi       = {10.18653/v1/2021.acl-long.393},
  bibsource = {dblp computer science bibliography, https://dblp.org}
}

@INPROCEEDINGS{predictingsemantic, 
author={B. {Xu} and D. {Ye} and Z. {Xing} and X. {Xia} and G. {Chen} and S. {Li}}, 
booktitle={2016 31st IEEE/ACM International Conference on Automated Software Engineering (ASE)}, 
title={Predicting semantically linkable knowledge in developer online forums via convolutional neural network}, 
year={2016}, 
pages={51-62}, 
}

@book{croft2010search,
  title={Search engines: Information retrieval in practice},
  author={Croft, W Bruce and Metzler, Donald and Strohman, Trevor},
  volume={283},
  year={2010},
  publisher={Addison-Wesley Reading}
}

@inproceedings{dong2015question,
  title={Question answering over freebase with multi-column convolutional neural networks},
  author={Dong, Li and Wei, Furu and Zhou, Ming and Xu, Ke},
  booktitle={Proceedings of the 53rd Annual Meeting of the Association for Computational Linguistics and the 7th International Joint Conference on Natural Language Processing (Volume 1: Long Papers)},
  volume={1},
  pages={260--269},
  year={2015}
}

@article{yih2015semantic,
  title={Semantic parsing via staged query graph generation: Question answering with knowledge base},
  author={Yih, Scott Wen-tau and Chang, Ming-Wei and He, Xiaodong and Gao, Jianfeng},
  year={2015}
}

@article{chen2017reading,
  title={Reading wikipedia to answer open-domain questions},
  author={Chen, Danqi and Fisch, Adam and Weston, Jason and Bordes, Antoine},
  journal={arXiv preprint arXiv:1704.00051},
  year={2017}
}

@article{miller2016key,
  title={Key-value memory networks for directly reading documents},
  author={Miller, Alexander and Fisch, Adam and Dodge, Jesse and Karimi, Amir-Hossein and Bordes, Antoine and Weston, Jason},
  journal={arXiv preprint arXiv:1606.03126},
  year={2016}
}

@article{rajpurkar2018know,
  title={Know What You Don't Know: Unanswerable Questions for SQuAD},
  author={Rajpurkar, Pranav and Jia, Robin and Liang, Percy},
  journal={arXiv preprint arXiv:1806.03822},
  year={2018}
}

@inproceedings{vinyals2015pointer,
  title={Pointer networks},
  author={Vinyals, Oriol and Fortunato, Meire and Jaitly, Navdeep},
  booktitle={Advances in Neural Information Processing Systems},
  pages={2692--2700},
  year={2015}
}

@inproceedings{wang2018multi,
  title={Multi-granularity hierarchical attention fusion networks for reading comprehension and question answering},
  author={Wang, Wei and Yan, Ming and Wu, Chen},
  booktitle={Proceedings of the 56th Annual Meeting of the Association for Computational Linguistics (Volume 1: Long Papers)},
  volume={1},
  pages={1705--1714},
  year={2018}
}

@article{devlin2018bert,
  title={Bert: Pre-training of deep bidirectional transformers for language understanding},
  author={Devlin, Jacob and Chang, Ming-Wei and Lee, Kenton and Toutanova, Kristina},
  journal={arXiv preprint arXiv:1810.04805},
  year={2018}
}

@inproceedings{vaswani2017attention,
  title={Attention is all you need},
  author={Vaswani, Ashish and Shazeer, Noam and Parmar, Niki and Uszkoreit, Jakob and Jones, Llion and Gomez, Aidan N and Kaiser, {\L}ukasz and Polosukhin, Illia},
  booktitle={Advances in Neural Information Processing Systems},
  pages={5998--6008},
  year={2017}
}

@article{ranzato2015sequence,
  title={Sequence level training with recurrent neural networks},
  author={Ranzato, Marc'Aurelio and Chopra, Sumit and Auli, Michael and Zaremba, Wojciech},
  journal={arXiv preprint arXiv:1511.06732},
  year={2015}
}

@article{yuan2017machine,
  title={Machine comprehension by text-to-text neural question generation},
  author={Yuan, Xingdi and Wang, Tong and Gulcehre, Caglar and Sordoni, Alessandro and Bachman, Philip and Subramanian, Sandeep and Zhang, Saizheng and Trischler, Adam},
  journal={arXiv preprint arXiv:1705.02012},
  year={2017}
}

@inproceedings{he2016dual,
  title={Dual learning for machine translation},
  author={He, Di and Xia, Yingce and Qin, Tao and Wang, Liwei and Yu, Nenghai and Liu, Tieyan and Ma, Wei-Ying},
  booktitle={Advances in Neural Information Processing Systems},
  pages={820--828},
  year={2016}
}

@inproceedings{xu2017answerbot,
  title={AnswerBot: automated generation of answer summary to developers{\'z} technical questions},
  author={Xu, Bowen and Xing, Zhenchang and Xia, Xin and Lo, David},
  booktitle={Proceedings of the 32nd IEEE/ACM International Conference on Automated Software Engineering},
  pages={706--716},
  year={2017},
  organization={IEEE Press}
}

@article{wang2018understanding,
  title={Understanding the factors for fast answers in technical Q\&A websites},
  author={Wang, Shaowei and Chen, Tse-Hsun and Hassan, Ahmed E},
  journal={Empirical Software Engineering},
  volume={23},
  number={3},
  pages={1552--1593},
  year={2018},
  publisher={Springer}
}

@inproceedings{carbonell1998use,
  title={The use of MMR, diversity-based reranking for reordering documents and producing summaries},
  author={Carbonell, Jaime and Goldstein, Jade},
  booktitle={Proceedings of the 21st annual international ACM SIGIR conference on Research and development in information retrieval},
  pages={335--336},
  year={1998},
  organization={ACM}
}

@article{LSA,
  title={Using latent semantic analysis in text summarization and summary evaluation},
  author={Steinberger, Josef and Jezek, Karel},
  journal={Proc. ISIM},
  volume={4},
  pages={93--100},
  year={2004}
}

@article{lexrank,
  title={Lexrank: Graph-based lexical centrality as salience in text summarization},
  author={Erkan, G{\"u}nes and Radev, Dragomir R},
  journal={Journal of artificial intelligence research},
  volume={22},
  pages={457--479},
  year={2004}
}

@inproceedings{textrank,
  title={Textrank: Bringing order into text},
  author={Mihalcea, Rada and Tarau, Paul},
  booktitle={Proceedings of the 2004 conference on empirical methods in natural language processing},
  year={2004}
}

@inproceedings{KL-sum,
  title={Exploring content models for multi-document summarization},
  author={Haghighi, Aria and Vanderwende, Lucy},
  booktitle={Proceedings of Human Language Technologies: The 2009 Annual Conference of the North American Chapter of the Association for Computational Linguistics},
  pages={362--370},
  year={2009},
  organization={Association for Computational Linguistics}
}

@inproceedings{BLEU,
  title={BLEU: a method for automatic evaluation of machine translation},
  author={Papineni, Kishore and Roukos, Salim and Ward, Todd and Zhu, Wei-Jing},
  booktitle={Proceedings of the 40th annual meeting on association for computational linguistics},
  pages={311--318},
  year={2002},
  organization={Association for Computational Linguistics}
}

@article{ROUGE,
  title={Rouge: A package for automatic evaluation of summaries},
  author={Lin, Chin-Yew},
  journal={Text Summarization Branches Out},
  year={2004}
}

@article{pagerank,
  title={Google’s pagerank},
  author={Wills, Rebecca S},
  journal={The Mathematical Intelligencer},
  volume={28},
  number={4},
  pages={6--11},
  year={2006},
  publisher={Springer}
}

@article{wordpiece,
  title={Google's neural machine translation system: Bridging the gap between human and machine translation},
  author={Wu, Yonghui and Schuster, Mike and Chen, Zhifeng and Le, Quoc V and Norouzi, Mohammad and Macherey, Wolfgang and Krikun, Maxim and Cao, Yuan and Gao, Qin and Macherey, Klaus and others},
  journal={arXiv preprint arXiv:1609.08144},
  year={2016}
}

@article{wang2017r-net,
  title={R-NET: Machine reading comprehension with self-matching networks},
  author={Wang, W and Yang, N and Wei, F and Chang, B and Zhou, M},
  journal={Natural Lang. Comput. Group, Microsoft Res. Asia, Beijing, China, Tech. Rep},
  volume={5},
  year={2017}
}

@Inbook{deeplearning-book,
  title={Deep learning},
  author={Goodfellow, Ian and Bengio, Yoshua and Courville, Aaron and Bengio, Yoshua},
  volume={1},
  year={2016},
  publisher={MIT press Cambridge}
}

@article{ye2017structure,
  title={The structure and dynamics of knowledge network in domain-specific q\&a sites: a case study of stack overflow},
  author={Ye, Deheng and Xing, Zhenchang and Kapre, Nachiket},
  journal={Empirical Software Engineering},
  volume={22},
  number={1},
  pages={375--406},
  year={2017},
  publisher={Springer}
}

@article{calefato2018ask,
  title={How to ask for technical help? Evidence-based guidelines for writing questions on Stack Overflow},
  author={Calefato, Fabio and Lanubile, Filippo and Novielli, Nicole},
  journal={Information and Software Technology},
  volume={94},
  pages={186--207},
  year={2018},
  publisher={Elsevier}
}

@article{chen2018data,
  title={Data-driven proactive policy assurance of post quality in community q\&a sites},
  author={Chen, Chunyang and Chen, Xi and Sun, Jiamou and Xing, Zhenchang and Li, Guoqiang},
  journal={Proceedings of the ACM on Human-Computer Interaction},
  volume={2},
  number={CSCW},
  pages={33},
  year={2018},
  publisher={ACM}
}

@inproceedings{zhou2018sequentialcopy,
  title={Sequential copying networks},
  author={Zhou, Qingyu and Yang, Nan and Wei, Furu and Zhou, Ming},
  booktitle={Thirty-Second AAAI Conference on Artificial Intelligence},
  year={2018}
}

@article{gu2016incorporatingcopy2seq2seq,
  title={Incorporating copying mechanism in sequence-to-sequence learning},
  author={Gu, Jiatao and Lu, Zhengdong and Li, Hang and Li, Victor OK},
  journal={arXiv preprint arXiv:1603.06393},
  year={2016}
}

@ARTICLE{massiveExplore-Britz:2017,
  author          = {{Britz}, D. and {Goldie}, A. and {Luong}, T. and {Le}, Q.},
  title           = "{Massive Exploration of Neural Machine Translation Architectures}",
  journal         = {ArXiv e-prints},
  archivePrefix   = "arXiv",
  eprinttype      = {arxiv},
  eprint          = {1703.03906},
  primaryClass    = "cs.CL",
  year            = 2017,
  month           = mar,
}

@inproceedings{6challenges-KoehnK17,
  author    = {Philipp Koehn and
               Rebecca Knowles},
  title     = {Six Challenges for Neural Machine Translation},
  booktitle = {Proceedings of the First Workshop on Neural Machine Translation, NMT@ACL
               2017, Vancouver, Canada, August 4, 2017},
  pages     = {28--39},
  year      = {2017},
}
